\documentclass[sigconf]{acmart}
\AtBeginDocument{%
  }

\copyrightyear{2026}
\acmYear{2026}
\setcopyright{cc}
\setcctype{by-nc-nd}
\acmConference[KDD '26]{Proceedings of the 32nd ACM SIGKDD Conference on Knowledge Discovery and Data Mining V.2}{August 09--13, 2026}{Jeju Island, Republic of Korea}
\acmBooktitle{Proceedings of the 32nd ACM SIGKDD Conference on Knowledge Discovery and Data Mining V.2 (KDD '26), August 09--13, 2026, Jeju Island, Republic of Korea}
\acmDOI{10.1145/3770855.3817748}
\acmISBN{979-8-4007-2259-2/2026/08}

\usepackage{hyperref}
\usepackage{url}
\usepackage{graphicx}
\usepackage{xspace}
\usepackage{makecell}
\usepackage{array}
\usepackage{algorithmicx,algorithm}
\usepackage{multirow}
\usepackage[normalem]{ulem}
\usepackage{subcaption}
\usepackage{tabularx}
\usepackage{color}
\usepackage[utf8]{inputenc}
\usepackage[T1]{fontenc}
\usepackage{CJKutf8}
\usepackage{enumitem}
\usepackage{diagbox}
\usepackage[flushleft]{threeparttable}
\usepackage{booktabs}
\usepackage{algorithmicx,algorithm}
\usepackage{algpseudocode}
\usepackage{wrapfig}
\usepackage[flushleft]{threeparttable}

\usepackage{amssymb}
\usepackage{subcaption}

\usepackage{pifont}

\begin{document}

\title{TiWeaver: Unified Temporal Dynamics Modeling via Contextual Patching}

\settopmatter{authorsperrow=3}
\author{Zhe Li}
\affiliation{
  \institution{East China Normal University}
  \country{Shanghai, China}
}
\email{zheli@stu.ecnu.edu.cn}

\author{Jindong Tian}
\affiliation{
  \institution{East China Normal University}
  \country{Shanghai, China}
}
\email{jdtian@stu.ecnu.edu.cn}

\author{Hao Miao}
\affiliation{
  \institution{Hong Kong Polytechnic University}
  \country{Hong Kong, China}
}
\email{hao.miao@polyu.edu.hk}

\author{Zhi Lei}
\affiliation{
  \institution{East China Normal University}
  \country{Shanghai, China}
}
\email{zhilei@stu.ecnu.edu.cn}

\author{Chenjuan Guo}
\affiliation{
  \institution{East China Normal University}
  \country{Shanghai, China}
}
\email{cjguo@dase.ecnu.edu.cn}

\author{Bin Yang}
\authornote{Corresponding author.}
\affiliation{
  \institution{East China Normal University}
 \country{Shanghai, China}
}
 \affiliation{
   \institution{Aalborg University}
  \country{Aalborg, Denmark}
 }
\email{byang@cs.aau.dk}

\renewcommand{\shortauthors}{Zhe Li et al.}


\begin{abstract}

Multivariate time series forecasting plays a critical role in real-world applications, including weather prediction, stock analysis, and health monitoring. Due to the diversity of data sources, time series exhibit diverse temporal dynamics, often accompanied by various irregularities such as missing values and non-uniform sampling frequencies. 
Such irregularities lead to complex and asynchronous temporal dependencies across channels. Thus, a single model with a fixed patching scheme often fails to adapt well to diverse multivariate time series, hindering accurate forecasting. 
In this paper, we propose TiWeaver, a unified framework designed to handle temporal dynamics and fine-grained inter-channel dependencies adaptively. 
Specifically, we introduce a Graph-Guided Adaptive Tokenizer (G$^2$AT) that divides time series into high contextually coherent patches by jointly considering temporal density and representation consistency. 
In addition, we propose a Fine-grained Asynchronous Dependency Extractor (FADE), which is designed to model fine-grained asynchronous inter-channel dependencies while incorporating long-term historical dependencies. 
We evaluate TiWeaver on 12 real-world time series datasets, where it achieves state-of-the-art performance, outperforming existing methods up to 25\%. These results demonstrate its robustness and effectiveness across diverse domains and data characteristics.

\end{abstract}



\maketitle

\newcommand\kddavailabilityurl{https://doi.org/10.5281/zenodo.20424563}

\ifdefempty{\kddavailabilityurl}{}{
\begingroup\small\noindent\raggedright\textbf{Resource Availability:}\\
The source code of this paper has been made publicly available at \url{\kddavailabilityurl}.
\endgroup
}

\section{Introduction}
\label{intro}
Multivariate time series (MTS) forecasting is widely used in numerous real-world applications, such as weather prediction~\cite{tian2025iclr,nips25Jiang,tian2025arrow}, stock analysis~\cite{acl23yu,cikm25mei}, health monitoring~\cite{nips25bi,liu2023large,lan2025gem}, transportation~\cite{xu2026most,guo2014trans,lu2011spatio,sb2023LightPath, ma2014,xu2025kdd}, and others\cite{li2026multi,sun2026beyond,li2024cikm,li2025TSFM-Bench}, playing a core role in optimizing operations and enhancing decision-making quality across various domains.

\begin{figure}[t]
  \centering
  \includegraphics[width=1\linewidth]{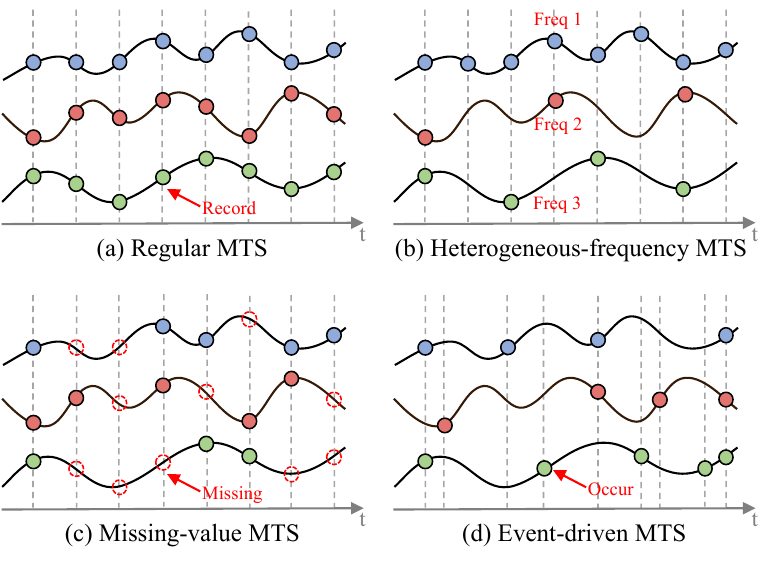}
  \caption{
  These four categories of multivariate time series (MTS) share identical underlying dynamics (indicated by the black line) yet differ in their observations.}
  \label{fig: data_type}
\end{figure}

MTS originate from diverse sources (e.g., fixed sensor networks, wearable devices, and industrial machinery), leading to heterogeneity in sampling strategies and inter-channel dependencies, often accompanied by varying irregularities. As illustrated in Figure~\ref{fig: data_type} and Table~\ref{tab:data_type}, we categorize MTS into four types: 
(a) \textbf{Regular MTS}, uniformly sampled across all channels, resulting in synchronous dependencies; 
(b) \textbf{Heterogeneous-frequency irregular MTS}, sampled at channel-specific rates, exhibiting asynchronous dependencies; 
(c) \textbf{Missing-value irregular MTS}, with absent observations due to packet loss or device failure, obtaining mixed synchronous and asynchronous dependencies; 
and (d) \textbf{Event-driven irregular MTS}, recorded randomly upon events, showing more predominantly asynchronous dependencies.

\begin{table}[h]
\centering
\caption{Sampling strategies and inter-channel dependencies across different MTS categories.} 
\resizebox{1\columnwidth}{!}{
\begin{tabular}{c|c|c}
\toprule
\textbf{Data Types} & \textbf{Sampling Strategies} & \textbf{Inter-channel Dependencies }\\
\midrule
\textbf{Regular MTS} & Consistent & Synchronous \\
\textbf{Heterogeneous-frequency MTS} & Multiple & Asynchronous \\
\textbf{Missing-value MTS} & Consistent\&Missing & Synchronous\&Asynchronous \\
\textbf{Event-driven MTS} & Random & Asynchronous\\
\bottomrule
\end{tabular}
}
\label{tab:data_type}
\end{table}

\begin{figure}[t]
  \centering
  \includegraphics[width=1\linewidth]{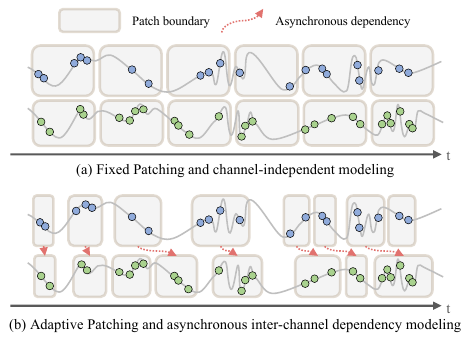}
  \caption{Comparison of patching strategies and channel-dependency modeling for MTS.}
  \label{fig: patching}
\end{figure}

However, existing methods based on fixed patching~\cite{patchtst, cirstea2022triformer} and channel-independent modeling~\cite{sun2026tmae, Wang2025CSformer} struggle to accommodate such heterogeneity in both aspects, motivating a unified approach for diverse MTS. Nonetheless, developing such a model entails two key challenges:

\textbf{Inflexible patching strategy for diverse MTS.}
MTS typically represent specific underlying dynamics that encompass temporal features such as periodicity and trends~\cite{timesnet}. Consequently, an effective patching strategy should tokenize MTS by considering the intrinsic dynamics. 
Existing methods typically apply fixed patching strategies uniformly across all channels, based on either fixed observation points or fixed time points~\cite{patchtst,liu2023itransformer,nips20cao}.
However, different channels do not necessarily share the same temporal features. Consequently, applying a uniform patching strategy across all channels may inadvertently distort the underlying dynamics themselves.
Moreover, fixed tokenization limits the adaptive extraction of temporal features, particularly for irregular MTS with incomplete observations. 
As shown in Figure~\ref{fig: patching}(a), fixed patches cut off the periodicity and trends of MTS, making it inappropriate for modeling temporal dependencies.
Although some recent studies begin to explore adaptive patching~\cite{mosaic,apn,lightgts,cheng2024darf}, they rely heavily on fixed and predefined settings, which fails to achieve flexible tokenization.
Given these limitations, one of the primary challenges in modeling MTS is designing a flexible patching strategy.

\textbf{Fine-Grained asynchronous inter-channel dependencies.}
Due to the limited observations of the underlying dynamics in MTS, especially in the latter three irregular MTS, temporal continuity is disrupted, which induces asynchronous dependencies as summarized in Table~\ref{tab:data_type}. More importantly, adaptive patching based on channel-specific dynamics further leads to fine-grained asynchronous inter-channel dependencies as illustrated in Figure~\ref{fig: patching}(b). 
However, many existing methods rely on channel-independent modeling~\cite{patchtst,qiu2026comprehensive}, which totally ignore the dependencies among channels, or construct inter-channel correlation graphs~\cite{TimeCHEAT,liu2026astgi,GraFITi}, which capture only coarse-grained, series-level dependencies and therefore largely overlook fine-grained asynchronous interactions. 
Such coarse-grained inter-channel dependency modeling typically assumes synchronous alignment and aggregates interactions over the entire series, making it difficult to capture fine-grained asynchronous inter-channel dependencies induced by heterogeneous temporal dynamics in diverse MTS. 
Therefore, with the dynamic-aware adaptive patching strategy in Figure~\ref{fig: patching}(b), it is necessary to explicitly capture these patch-level dependencies.

\begin{figure*}[t]
  \centering
  \includegraphics[width=1\linewidth]{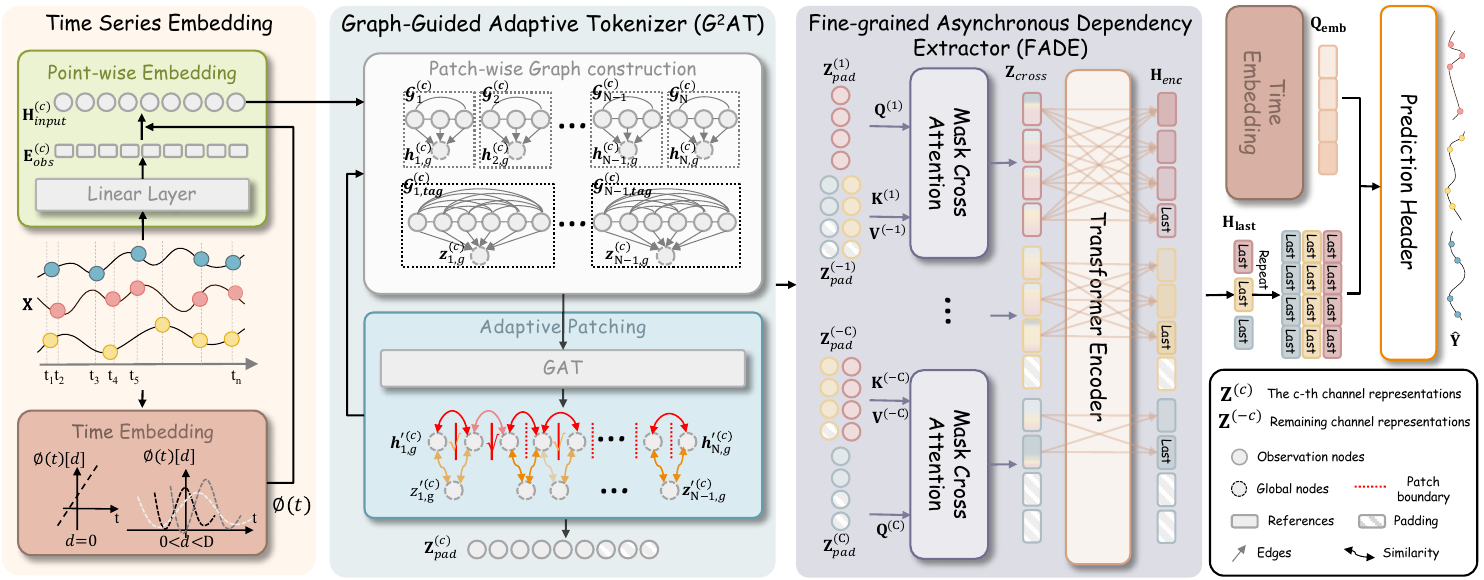}
  \caption{The overall framework of TiWeaver. On the left, a time series sample is mapped into a high-dimensional latent space through time embedding and point-wise embedding. The \textbf{G$^2$AT} adaptively divides series into high contextually coherent patches based on temporal dynamics. The \textbf{FADE} captures asynchronous inter-channel dependencies and long-term historical dependencies across patches. Finally, on the far right, the Prediction Header generates the final forecasts based on time queries.}
  \label{fig: model}
\end{figure*}

In this paper, we propose TiWeaver, a unified temporal dynamics modeling for multivariate time series.
To address \textbf{the first challenge}, we design a Graph-Guided Adaptive Tokenizer (G$^2$AT) that adaptively partitions diverse MTS into contextually coherent patches, implemented through a bottom-up aggregation process. 
Specifically, G$^2$AT first splits MTS into minimal patch units (e.g., individual time points) and employs a Graph Attention Network (GAT)~\cite{gat} to capture local temporal dependencies, yielding patch-level representations. Then G$^2$AT merges adjacent patches based on their temporal density and representation consistency to maintain high contextual coherence within each resulting patch. This procedure is repeated iteratively until the model learns clear patches boundaries, ensuring contextual separability among them. 
Therefore, for any MTS, G$^2$AT can learn locally dynamics-consistent representations that reflect the underlying dynamics of each channel, thereby providing a unified tokenization approach.

These representations exhibit complex fine-grained asynchronous inter-channel dependencies. To address \textbf{the second challenge}, we introduce a Fine-Grained Asynchronous Dependency Extractor (FADE). This module is intended to capture fine-grained asynchronous dependencies, modeling both inter-channel correlations and intra-channel temporal correlations across patches. Specifically, FADE treats patch-level representations of each channel as queries and those of other channels as keys/values, and applies masked cross-attention to model asynchronous inter-channel dependencies (e.g., lagged cross-variable effects). Meanwhile, since time series often exhibit long-term dependencies across patches, we further adopt a Transformer-based encoder-only structure to model these temporal dependencies. Ultimately, the last patch representation of each channel integrates local temporal dependencies, asynchronous inter-channel dependencies and long-term historical dependencies, enabling FADE to explicitly capture complex cross-channel interactions from a fine-grained perspective.

In summary, our contributions can be summarized as follows:
\begin{itemize}[noitemsep, topsep=0pt, leftmargin=*]

  \item We design a novel Graph-Guided Adaptive Tokenizer (G$^2$AT) that iteratively divides the whole sequence into contextually coherent patches, providing a unified tokenization scheme for MTS with diverse temporal dynamics. 
  \item We introduce a Fine-Grained Asynchronous Dependency Extractor (FADE) to capture fine-grained asynchronous inter-channel dependencies while enhancing the modeling of long-term historical interactions.
  \item Extensive experiments on 12 real-world datasets spanning 6 domains demonstrate the robustness and effectiveness of TiWeaver.
\end{itemize}

\section{Preliminaries}

Multivariate time series forecasting learns a mapping from historical observations to future values. We provide unified definitions for the data and the forecasting task:

\noindent \textbf{Definition (Multivariate Time Series, MTS).}
A MTS sample $\mathbf{X}= \langle( \boldsymbol{x}_t, t, \boldsymbol{m}_t)\rangle_{t=1}^T$ is a sequence of triples. Here $\boldsymbol{x}_t \in \mathbb{R}^C$ denotes the observed value of $C$ channels at timestamp $t$, and $ \boldsymbol{m}_t \in \{0,1\}^C $ is a binary mask vector, where $m_t^{(c)}$ indicates whether $c$-th channel is observed at $t$.

\noindent \textbf{Problem (Multivariate Time Series Forecasting).}
For diverse MTS, given historical observations $\mathbf{X}$, the goal is to forecast future sequence $\hat{\mathbf{Y}} = \langle\hat{\boldsymbol{x}}_{t}\rangle_{t=T+1}^{T+F}$ at future query timestamps $\mathbf{Q} = \langle q_t \rangle_{t=T+1}^{T+F}$, formalizing as $\mathcal{F}(\mathbf{X}, \mathbf{Q}) \rightarrow \hat{\mathbf{Y}}$, where $\mathcal{F}(\cdot)$ denotes the forecasting model to be learned and $F$ is the forecast horizon.

\section{Methodology}

\subsection{Model Overview}
We propose TiWeaver, a unified temporal dynamics modeling method via contextual patching. The overall model architecture is illustrated in Figure~\ref{fig: model}, including four core modules: Time Series Embedding, Graph-Guided Adaptive Tokenizer (G$^2$AT), Fine-Grained Asynchronous Dependency Extractor (FADE) and Prediction Header.

\textbf{First}, TiWeaver employs dual-branch embeddings to learn trainable representations for timestamps and observed values, and fuse them into point-level representations, enhancing TiWeaver’s sensitivity to temporal dynamics (Section~\ref{time}). \textbf{Second}, these representations are processed by Graph-Guided Adaptive Tokenizer (G$^2$AT) (Section~\ref{gap}). Specifically, we divide MTS into minimal units to construct patch-level graphs and apply a Graph Attention Network (GAT) to capture intra-patch local temporal dependencies, producing patch-level representations. G$^2$AT further learns clear patch boundaries adaptively based on the temporal density and representation consistency between adjacent patches, ensuring high contextual coherence within each patch while maintaining sufficient separation across patches. \textbf{Third}, the locally dynamic-consistent representations can be obtained, we introduce a masked cross-variable attention mechanism to capture asynchronous inter-channel dependencies and employ an encoder-only architecture to model long-term historical dependencies (see Section~\ref{cvi}). \textbf{Finally}, the final representations, together with the target forecasting timestamps, are fed into a Feed-Forward Network (FFN) to generate the forecasts (see Section~\ref{Forecasting}).

\subsection{Time Series Embedding}
\label{time}

Multivariate time series often exhibit diverse irregularities, producing non-uniform timestamps. An explicit encoding timestamp is essential for effective representation learning across diverse MTS. 
Learnable time embeddings~\citep{timeemb} is utilized to encode the timestamps, enabling the model to perceive fine-grained time intervals. Given a timestamp $t$, its embedding $\phi(t)$ is defined as:
\begin{equation}
\text{TimeEmb}(t) = \phi(t) = 
\begin{cases}
\omega_d \cdot t + \beta_d, & \text{if } d = 0 \\
\sin(\omega_d \cdot t + \beta_d), & \text{if } 0 < d < D
\end{cases} \ ,
\end{equation}
where $\omega_d$ and $\beta_d$ are learnable parameters and $D$ is the embedding dimension. This formula encodes timestamps into a high-dimensional space, where the linear term captures non-periodic trends as time progresses, and the sinusoidal terms capture temporal periodicity.

For the observations in each sample $\mathbf{X}$, we apply a linear layer to obtain the value embeddings $\mathbf{E}_{\text{obs}}=\langle\boldsymbol{e}_t\rangle_{t=1}^T$. 
We then add each channel point-wise embedding $\boldsymbol{e}_t^{(c)} \in \mathbb{R}^{C \times D}$ to its time embedding to enrich temporal context:
\begin{equation}
\boldsymbol{h}_{t}^{(c)} = \phi(t) + \boldsymbol{e}_{t}^{(c)} \in \mathbb{R}^D.
\end{equation}
If a channel is unobserved at $t$, we set $\boldsymbol{h}_t^{(c)}=\boldsymbol{0}$ with a mask $m_t^{(c)}=0$, which will be ignore in the following steps. This process yields the input embedding representation $\mathbf{H}_{input} \in \mathbb{R}^{T \times C \times D}$.

\subsection{Graph-Guided Adaptive Tokenizer}
\label{gap}

Multivariate time series forecasting aims to capture the temporal dependencies across time steps. 
However, modeling time points alone is insufficient to capture context-aware temporal dependencies, and fixed patch sizes often fail to adapt to complex temporal dynamics.
Therefore, an adaptive patching strategy is essential for establishing a unified tokenization paradigm for diverse MTS. In this section, we propose a Graph-Guided Adaptive Tokenizer (G$^2$AT) that divides MTS into contextually coherent patches. 
Our approach begins with unit patches (e.g., individual time points) and merges them via a bottom-up process.

\subsubsection{Patch-wise Graph Construction}
To capture local temporal dependencies, we apply GAT~\cite{gat} over each patch-level graph. Specifically, given the $c$-th channel sequence $\mathbf{H}_{input}^{(c)}=\langle\boldsymbol{h}_{t}^{(c)}\rangle_{t=1}^T$ and a predefined minimum observation count $P_{min}$, we first partition the sequence into $N = \frac{T}{P_{min}}$ non-overlapping patches, each patch having $P_{min}$ observations. 

The $n$-th patch, whether from the initial patching segmentation or adaptive patching, is modeled as a directed graph $\mathcal{G}_n^{(c)}=\{\mathcal{V}_n^{(c)}, \mathcal{E}_n^{(c)}\}$ (Ref. to the gray dashed rectangle in Figure~\ref{fig: model}). The node set $\mathcal{V}_n^{(c)}$ consists of observation nodes $v_{n,i}^{(c)}$ (Ref. to solid circles; initialized as $\boldsymbol{h}^{(c)}_{n,i} = \boldsymbol{h}^{(c)}_{t}$, where $t = (n-1)P_{min} + i$) and a virtual global node $v^{(c)}_{n,glob}$ (Ref. to dashed circle; initialized as $\mathbf{0}$) designed to aggregate patch-level information. To capture dependencies and control information flow, the edge set $\mathcal{E}_n^{(c)}$ is constructed by fully connecting all observation nodes bidirectionally (Ref. to solid lines), alongside directed edges pointing from each observation node to the global node (Ref. to solid arrows). Based on this topology, for any node $v^{(c)}_{n,i} \in \mathcal{V}_n^{(c)}$, the attention score with its neighbor $v^{(c)}_{n,j}$ is computed as follows:
\begin{equation}
e_{n,ij}^{(c)} = \text{LeakyReLU} \left( \boldsymbol{a}_{n,o}^{(c)\top} \left[ \mathbf{W}_{n,o} \boldsymbol{h}_{n,i}^{(c)} \,\|\, \mathbf{W}_{n,o} \boldsymbol{h}_{n,j}^{(c)} \right] \right)
\label{eq4}
\end{equation}
where $\mathbf{W}_{n,o} \in \mathbb{R}^{D \times D}$ is a linear projection and $\boldsymbol{a}_{n,o} \in \mathbb{R}^{2D}$ is a learnable attention vector. The attention weights are normalized across all neighbors $\mathcal{N}_{n,i}$ of $v_{n,i}$:
\begin{equation}
\alpha_{n,ij}^{(c)} = \frac{ \exp(e_{n,ij}^{(c)}) }{ \sum_{k \in \mathcal{N}_{n,i}} \exp(e_{n,ik}^{(c)}) },
\label{eq5}
\end{equation}
The representation of node $v_{n,i}$ is updated as follows:
\begin{equation}
\boldsymbol{h}'^{(c)}_{n,i} = \sigma \left( \sum_{j=1}^{\mathcal{N}}\alpha_{n,ij}^{(c)} \mathbf{W}_{n,o} \boldsymbol{h}^{(c)}_{n,j} \right),
\label{eq6}
\end{equation}
where $ \sigma(\cdot) $ denotes an activation function. The global node representation of the $n$-th patch is noted as $\boldsymbol{h}_{n,g}'^{(c)}$.

\subsubsection{Adaptive Patching}

To capture the temporal dynamics of diverse MTS and obtain locally dynamic-consistent representations, an adaptive patching strategy is designed to learn clear boundaries among minimal patch units. The overall strategy is summarized in Algorithm~\ref{gapm}, and we describe it step-by-step below.

To measure the consistency of representations between adjacent patches $\boldsymbol{h}_{n,g}'^{(c)}$ and $\boldsymbol{h}_{n+1,g}'^{(c)}$, we compute their similarity $S_n$ as follows (Ref. to the red bidirectional arrow in Figure~\ref{fig: model}):
\begin{equation}
S_n^{(c)}=\text{Sim}(\boldsymbol{h}_{n,g}'^{(c)}, \boldsymbol{h}_{n+1,g}'^{(c)}) = \frac{ \boldsymbol{h}_{n,g}'^{(c)\top} \boldsymbol{h}_{n+1,g}'^{(c)} }{ \| \boldsymbol{h}_{n,g}'^{(c)} \|_2 \cdot \| \boldsymbol{h}_{n+1,g}'^{(c)} \|_2 },
\label{eq7}
\end{equation}
the operator $\| \cdot \|_2$ represents the Euclidean norm (L2-norm). 

Due to the complex irregularities of diverse MTS, temporal continuity is interrupted. Therefore, we define the ``temporal density'' between adjacent patches to explicitly weight inter-patch dependencies according to their time intervals. Specifically, for the $n$-th patch with start timestamps $t_{n,s}^{(c)}$ and end timestamps $t_{n,e}^{(c)}$, $\Delta{t}_n^{(c)}=(t_{n+1,s}^{(c)}-t_{n,e}^{(c)})$ is the time intervals between adjacent patches. Based on $\Delta{t}_n$, temporal density $\delta_{n}$ can be calculated as follows:
\begin{equation}
\delta_n^{(c)} = \exp\left(-\frac{\Delta t_n}{\max\limits_{1\leq k\leq N'-1} \Delta t_k}\right)\in (0, 1],
\label{eq8}
\end{equation}
where $max(\cdot)$ returns the maximum time interval among all patch pairs, and $N'$ denotes the number of patches generated during the adaptive patching process, which varies across different variables. Larger time intervals correspond to lower temporal density.

Adaptive patching process is initialized with an all-zero boundary indicator, denoted as $\mathcal{B}^{(c)} = \langle b_n^{(c)} = 0\rangle_{n=1}^{N-1}$ (Ref. to the red dashed line in Figure~\ref{fig: model}), indicating that input time series is initially treated as an indivisible sequence. During the bottom-up process, each element $b_n^{(c)}$ is updated based on the contextually coherent between adjacent patches, quantified by their temporal density $\delta_n$ and representation similarity $S_n$:
\begin{equation}
b_n^{(c)} = 
\begin{cases}
1, & \text{if } S_n^{(c)} * \delta_n^{(c)} < \tau \\
0, & \text{otherwise}
\end{cases} \ .
\label{eq9}
\end{equation}
where $\tau$ is a predefined coherent threshold and $b_n^{(c)}=1$ indicates that the two adjacent patches should not be treated entirely (Ref. to the red check mark with red solid line in Figure~\ref{fig: model}).

Unfortunately, for adjacent patches whose boundary remains 0, we cannot assume that they are highly contextually coherent, even if their representations are similar and their temporal interval is small. This is because different underlying mechanisms can produce similar local patterns (e.g., repeated periodic fragments), while the whole sequence may still exhibit short-term abrupt changes, resulting in non-smooth dynamics. Therefore, we treat all adjacent patches as contextually coherent candidates and construct new graphs $\{\mathcal{G}_{n,tag}^{(c)}\}_{n=1}^{N-1}$ (Ref. to the black dashed rectangle in Figure~\ref{fig: model}). By computing the similarity between each candidate and its original patches, we encourage smooth dynamics between adjacent patches and update the boundary $\mathcal{B}^{(c)}$ (Ref. to the orange check mark with red solid line in Figure~\ref{fig: model}):
\begin{equation}
    b_{n}^{(c)} = 
    \begin{cases}
    1, & \text{if } S_{n,n}^{(c)} < \tau \ or \ S_{n,n+1}^{(c)} < \tau \\
    0, & \text{otherwise}
    \end{cases},
    \label{eq10}
\end{equation}
where $S_{n,n}^{(c)}$ denotes the similarity between the $n$-th candidate $\mathbf{Z}_{n}^{(c)}$ and its original $i$-th patch $\mathbf{H}_{n}^{(c)}$. Notably, since these candidates overlap, time density is not considered. We iteratively perform the above procedure until boundaries are established between all adjacent patches (i.e., $\mathcal{B}^{(c)} = \{\mathbf{1}\}$).

\begin{algorithm}[t]
\caption{Adaptive Patching}
\small
\begin{flushleft}
{\bf Input:} 
The $c$-th channel time series $\mathbf{H}_{input}^{(c)}\in \mathbb{R}^{T\times D}$

{\bf Output:}
The patches representations $\mathbf{Z}_{final}^{(c)}\in \mathbb{R}^{N_c'\times D}$

\end{flushleft}
\begin{algorithmic}[1]
\State Divide $\mathbf{H}_{input}^{(c)}$ into patches with $P_{min}$;
\State Initialize intra-patch graphs $\mathcal{G}_n^{(c)} =\{\mathcal{V}_n^{(c)}, \mathcal{E}_n^{(c)}\}$;
\State Initialize boundary indicator $\mathcal{B}^{(c)} = \{b_n^{(c)} = 0\}_{n=1}^{N-1}$;
\State {\bf while} $\exists\ b_n^{(c)} \in \mathcal{B}^{(c)} \text{ s.t. } b_n^{(c)} = 0$ {\bf do}
\State \hspace{0.1in} Obtain patch-level representations $\boldsymbol{h}_{n,g}'^{(c)}$ using Eqs.~\eqref{eq4}–\eqref{eq6};
\State \hspace{0.1in} Get similarity of adjacent patches $\mathbf{S}^{(c)}=\{S_n^{(c)}\}_{n=1}^{N-1}$ by Eq.~\eqref{eq7};
\State \hspace{0.1in} Compute time density $\mathbf{\delta}^{(c)} \in (0, 1]^{N-1}$ using Eq.~\eqref{eq8};
\State \hspace{0.1in} {\bf if} $\mathbf{S}^{(c)} \cdot \mathbf{\delta}^{(c)} < \tau$ {\bf then}
\State \hspace{0.2in} Update the boundary $b_n^{(c)} \gets 1$ in $\mathcal{B}^{(c)}$ as Eq.~\eqref{eq9};
\State \hspace{0.1in} {\bf end if}
\State \hspace{0.1in} Re-divide $\mathbf{H}_{input}^{(c)}$ into candidates $\mathbf{Z}^{(c)}$;
\State \hspace{0.1in} Reconstruct new patch-level graphs $\mathcal{G}_{n,tag}^{(c)}$;
\State \hspace{0.1in} Get new patch-level representations $\boldsymbol{z}_{n,g}'^{(c)}$ from Eqs.~\eqref{eq4}--\eqref{eq6};
\State \hspace{0.1in} Compute similarity $\{{(S_{n,n}^{(c)},S_{n,n+1}^{(c)})}\}_{i=1}^{N-2}$ as Eq.~\eqref{eq7};
\State \hspace{0.1in} {\bf if} $S_{n,n}^{(c)} < \tau$ {\bf or} $S_{n,n+1}^{(c)} < \tau$ {\bf then}
\State \hspace{0.2in} Update the boundary $b_n^{(c)} \gets 1$ in $\mathcal{B}^{(c)}$ using Eq.~\eqref{eq10};
\State \hspace{0.1in} {\bf end if}
\State \hspace{0.1in} Update $\mathcal{G}_n^{(c)}$;
\State {\bf end while}
\State {\bf return} Final patch representations $\mathbf{Z}_{final}^{(c)}$;
\end{algorithmic}
\label{gapm}
\end{algorithm} 

This procedure aligns with the Minimum Description Length (MDL) principle: it minimizes description length by using fewer patch representations while preserving information fidelity. We provide a theoretical analysis in Appendix~\ref{theoretical analysis} to substantiate that adaptive patching can model temporal dynamics more effectively than fixed patching, by maximizing intra-patch contextual coherence and promoting inter-patch contextual separability.

Due to diverse temporal dynamics across channels, G$^2$AT yields a channel-specific number $N'_c$ of representations $\mathbf{Z}_{final}^{(c)} \in \mathbb{R}^{N_c' \times D}$. We zero-pad each channel to the maximum length $N_{\max} = \max\limits_{1 \le c \le C} N'_c$ (Ref. to the gray solid-outlined circle with striped pattern in Figure~\ref{fig: model}) for batch-parallel processing, yielding $\mathbf{Z}_{pad} \in \mathbb{R}^{C \times N_{\text{max}} \times D}$. 
A binary validity mask $ \mathbf{M} \in \{0, 1\}^{C \times N_{max}} $ indicates the padding positions:
\begin{equation}
\mathbf{M}[c][i] = 
\begin{cases}
1 & \text{if } i \leq N'_c  \  \\
0 & \text{otherwise}
\end{cases}.
\end{equation}

\subsection{Fine-Grained Asynchronous Dependency Extractor}
\label{cvi}

In MTS forecasting, the current state of a channel needs to consider earlier impact in itself and others (e.g., cumulative effects or delayed effects). Fine-Grained Asynchronous Dependency Extractor (FADE) provides a mechanism to retrieve contextual information from the right positions over a longer lookback window.

\subsubsection{Inter-Channel Interactions}

As illustrated in Figure~\ref{fig: patching}(b), due to diverse dynamics across channels, G$^2$AT produces a channel-specific number of patches. We introduce a masked cross-channel attention module to capture fine-grained asynchronous dependencies across channels while preventing padded patches from contributing to attention computation.
Formally, given a query channel representation $ \mathbf{Z}_{pad}^{(c)} \in \mathbb{R}^{N_{max} \times D} $ and representations of remaining channels, denotes as $ \mathbf{Z}_{pad}^{(-c)} \in \mathbb{R}^{(C-1)\cdot N_{max} \times D} $, cross-channel attention is defined as:
\begin{equation}
    \text{Cross-Attn}(\mathbf{Z}_{pad}^{(c)}, \mathbf{Z}_{pad}^{(-c)}) = \text{Softmax}\left( \frac{\mathbf{Q}^{(c)} \mathbf{K}^{(-c)\top}}{\sqrt{D}} + \mathbf{M}' \right) \mathbf{V}^{(-c)},
\end{equation}
where $\mathbf{Q}^{(c)} = \mathbf{Z}_{pad}^{(c)} \mathbf{W}_Q$, $\mathbf{K}^{(-c)} = \mathbf{Z}_{pad}^{(-c)} \mathbf{W}_K$, and $\mathbf{V}^{(-c)} = \mathbf{Z}_{pad}^{(-c)} \mathbf{W}_V$ are the linearly projected queries, keys, and values, respectively. The learnable projection matrices $\mathbf{W}_Q, \mathbf{W}_K, \mathbf{W}_V \in \mathbb{R}^{D \times D}$ are shared across channels and $\mathbf{M}'={\mathbf{M}^{(c)}(\mathbf{M}^{(-c)})}^{T} \in \mathbb{R}^{(C-1)\cdot N_{max}}$.

This design allows TiWeaver to dynamically capture asynchronous dependencies across channels and preserve computational efficiency, yielding multivariate representations $\mathbf{Z}_{cross} \in \mathbb{R}^{C \times N_{max} \times D}$.

\begin{table*}[t]
\small
\centering
\caption{Overall performance evaluated by MAE (mean ± std). The best-performing results are highlighted in bold, the second-best results are highlighted in underline. `$\dag$' represents the heterogeneous-frequency datasets obtained by downsampling different channels at different ratios. `$*$' denotes that the missing-value datasets with 20\% missing ratio.}
\resizebox{1\linewidth}{!}{
\begin{tabular}{c|ccc|ccc}
\toprule
\multirow{2}{*}{\textbf{Datasets}} & \multicolumn{3}{c|}{\textbf{Regular MTS}} & \multicolumn{3}{c}{\textbf{Heterogeneous-frequency MTS}} \\ \cmidrule{2-7} 
 & \textbf{Exchange} & \textbf{Weather} & \textbf{ZafNoo} & \textbf{Exchange$^\dag$} & \textbf{Weather$^\dag$} & \textbf{ZafNoo$^\dag$} \\ \midrule
NeuralFlows~\cite{nFlows} & 1.0726$\pm$0.1402 & 0.2599$\pm$0.0091 & 0.5277$\pm$0.0163 & 0.8157$\pm$0.0124 & 0.2487$\pm$0.0136 & 0.6634$\pm$0.0020 \\
GNeuralFlow~\cite{gnflows} & 1.1379$\pm$0.2716 & 0.5202$\pm$0.1205 & 0.6466$\pm$0.0486 & 1.0835$\pm$0.0506 & 0.5691$\pm$0.1078 & 0.7201$\pm$0.0319 \\
tPatchGNN~\cite{tpatchgnn} & 0.4329$\pm$0.0723 & 0.2387$\pm$0.0053 & 0.4685$\pm$0.0106 & 0.3450$\pm$0.0191 & 0.2354$\pm$0.0014 & 0.6225$\pm$0.0342 \\
GraFITi~\cite{GraFITi} & 0.4370$\pm$0.0284 & 0.2654$\pm$0.0175 & 0.4610$\pm$0.0051 & 0.3227$\pm$0.0485 & 0.2409$\pm$0.0093 & 0.6410$\pm$0.0060 \\
Hi-Patch~\cite{luo2025hipatch} & 0.5560$\pm$0.1093 & 0.2864$\pm$0.0076 & 0.5083$\pm$0.0231 & 0.3437$\pm$0.0168 & 0.2546$\pm$0.0119 & 0.6276$\pm$0.0340 \\
mTAND~\cite{mtan} & 0.8297$\pm$0.0452 & 0.3267$\pm$0.0045 & 0.4952$\pm$0.0138 & 0.8166$\pm$0.0371 & 0.2642$\pm$0.0192 & 0.4653$\pm$0.0071 \\
PrimeNet~\cite{primenet} & 1.4605$\pm$0.0061 & 0.6071$\pm$0.0000 & 0.7211$\pm$0.0018 & 1.4059$\pm$0.0003 & 0.5975$\pm$0.0000 & 0.7295$\pm$0.0001 \\ 
APN~\cite{apn} & 0.3491$\pm$0.0176 & 0.2626$\pm$0.2001 & 0.6373$\pm$0.1510 & 0.2703$\pm$0.0022 & 0.2367$\pm$0.0711 & 0.6446$\pm$0.0053 \\ \midrule
FEDformer~\cite{fedformer} & 0.2892$\pm$0.0051 & 0.2974$\pm$0.0085 & 0.6507$\pm$0.0624 & 0.7897$\pm$0.3247 & 0.3573$\pm$0.0488 & 0.6295$\pm$0.0845 \\
PatchTST~\cite{patchtst} & \uline{0.2030$\pm$0.0004} & 0.2131$\pm$0.0010 & 0.4188$\pm$0.0012 & 0.2273$\pm$0.0031 & \uline{0.2197$\pm$0.0010} & \uline{0.4490$\pm$0.0019} \\
iTransformer~\cite{liu2023itransformer} & 0.2304$\pm$0.0033 & 0.2510$\pm$0.0016 & 0.4245$\pm$0.0018 & 0.2725$\pm$0.0093 & 0.2875$\pm$0.0094 & 0.6618$\pm$0.0023 \\
TimesNet~\cite{timesnet} & 0.2519$\pm$0.0085 & 0.2292$\pm$0.0075 & 0.5484$\pm$0.0980 & 0.3542$\pm$0.0143 & 0.2375$\pm$0.0132 & 0.5760$\pm$0.1288 \\
PDF~\cite{dai2024periodicity} & 0.2057$\pm$0.0029 & \uline{0.2115$\pm$0.0009} & \textbf{0.4101$\pm$0.0019} & \uline{0.2249$\pm$0.0107} & 0.2201$\pm$0.0029 & 0.5071$\pm$0.0027 \\ 
TimeMosaic~\cite{mosaic} & 0.2098$\pm$0.3107 & 0.2209$\pm$0.0003 & 0.4185$\pm$0.0064 & 0.2161$\pm$0.0373 & 0.2561$\pm$0.0041 & 0.4432$\pm$0.0055  \\ \midrule
\textbf{TiWeaver} & \textbf{0.2005$\pm$0.0033} & \textbf{0.2089$\pm$0.0004} & \uline{0.4173$\pm$0.0017} & \textbf{0.1915$\pm$0.0011} & \textbf{0.2038$\pm$0.0041} & \textbf{0.4218$\pm$0.0036} \\

\bottomrule
\toprule

\multirow{2}{*}{\textbf{Datasets}} & \multicolumn{3}{c|}{\textbf{Missing-value MTS}} & \multicolumn{3}{c}{\textbf{Event-driven MTS}} \\ 
\cmidrule{2-7}
 & \textbf{Exchange$^*$} & \textbf{Weather$^*$}  & \textbf{ZafNoo$^*$} & \textbf{Human Activity} & \textbf{USHCN} & \textbf{PhysioNet2012} \\
\midrule
NeuralFlows~\cite{nFlows} & 0.8769$\pm$0.1042 & 0.2646$\pm$0.0015 & 0.6625$\pm$0.0058 & 0.3719$\pm$0.0132 & 0.3408$\pm$0.0038 & 0.4978$\pm$0.0025 \\
GNeuralFlow~\cite{gnflows} & 1.2046$\pm$0.0575 & 0.5401$\pm$0.0498 & 0.7186$\pm$0.0186 & 0.4382$\pm$0.0048 & 0.4700$\pm$0.1199 & 0.6569$\pm$0.0190 \\
tPatchGNN~\cite{tpatchgnn} & 0.3702$\pm$0.0026 & 0.2357$\pm$0.0023 & 0.6393$\pm$0.0178 & 0.1453$\pm$0.0014 & 0.3303$\pm$0.0226 & 0.5341$\pm$0.1673 \\
GraFITi~\cite{GraFITi} & 0.4365$\pm$0.0326 & 0.2582$\pm$0.0116 & 0.6302$\pm$0.0045 & 0.1745$\pm$0.0029 & \uline{0.3277$\pm$0.0166} & 0.5293$\pm$0.0005 \\
Hi-Patch~\cite{luo2025hipatch} & 0.4410$\pm$0.1070 & 0.2648$\pm$0.0028 & 0.6500$\pm$0.0150 & 0.1435$\pm$0.0013 & 0.3382$\pm$0.0192 & 0.4678$\pm$0.0004 \\
mTAND~\cite{mtan} & 0.7871$\pm$0.0677 & 0.2750$\pm$0.0061 & 0.4874$\pm$0.0192 & 0.2430$\pm$0.0105 & 0.6392$\pm$0.0706 & 0.4900$\pm$0.0006 \\
PrimeNet~\cite{primenet} & 1.4528$\pm$0.0004 & 0.6070$\pm$0.0001 & 0.7215$\pm$0.0001 & 1.7262$\pm$0.0005 & 0.5116$\pm$0.0004 & 0.6851$\pm$0.0000 \\  
APN~\cite{apn} & 0.3250$\pm$0.0092 & 0.2439$\pm$0.0330 & 0.6396$\pm$0.2012 & \uline{0.1421$\pm$0.0067} & 0.3637$\pm$0.0008 & \uline{0.4621$\pm$0.0035} \\ \midrule
FEDformer~\cite{fedformer} & 0.4070$\pm$0.0362 & 0.3490$\pm$0.0158 & 0.6365$\pm$0.0859 & 0.3606$\pm$0.0049 & 0.5218$\pm$0.1272 & 0.5126$\pm$0.0002 \\ 
PatchTST~\cite{patchtst} & 0.3227$\pm$0.0064 & 0.2398$\pm$0.0020 & \uline{0.4611$\pm$0.0013} & 0.1838$\pm$0.0004 & 0.4370$\pm$0.0716 & 0.5075$\pm$0.0200 \\
iTransformer~\cite{liu2023itransformer} & 0.3258$\pm$0.0155 & 0.2842$\pm$0.0062 & 0.6478$\pm$0.0206 & 0.2432$\pm$0.0083 & 0.4401$\pm$0.0193 & 0.5156$\pm$0.0018 \\
TimesNet~\cite{timesnet} & 0.3619$\pm$0.0487 & 0.2470$\pm$0.0022 & 0.5583$\pm$0.0011 & 0.2396$\pm$0.0010 & 0.3591$\pm$0.0003 & 0.5101$\pm$0.0013 \\
PDF~\cite{dai2024periodicity} & 0.2793$\pm$0.0087 & \uline{0.2317$\pm$0.0017} & 0.5507$\pm$0.0051 & - & - & - \\
TimeMosaic~\cite{mosaic} & \uline{0.2503$\pm$0.0411} & 0.3180$\pm$0.0050 & 0.4761$\pm$0.0032 & 0.2527$\pm$0.0129 & 0.3631$\pm$0.0061 & 0.5176$\pm$0.0603 \\ \midrule
\textbf{TiWeaver} & \textbf{0.2091$\pm$0.0061} & \textbf{0.2114$\pm$0.0078} & \textbf{0.4525$\pm$0.0024} & \textbf{0.1362$\pm$0.0006} & \textbf{0.2927$\pm$0.0013} & \textbf{0.4617$\pm$0.0002} \\ 
\bottomrule
\end{tabular}
}

\begin{tablenotes}
\fontsize{8}{2}
\selectfont
\item \textit{\textbf{Note:}} PDF imposes a rigid constraint between the input sequence length and the patch size, which are unable to handle event-driven datasets with variable-length inputs, as demonstrated in “–”.
\end{tablenotes}

\label{tab:main_result}
\end{table*}

\subsubsection{Intra-Channel Encoder}

To capture long-term historical dependencies, we feed $\mathbf{Z}_{cross}$ into a Transformer-based encoder-only module.
Similar to a standard Transformer encoder, the module comprises positional encoding, self-attention layers, layer normalization, and a feed-forward network. Notably, due to the presence of invalid padding tokens produced by G$^2$AT, the self-attention layer is also equipped with the validity mask $\mathbf{M}$: 
\begin{equation}
\text{Self-Attn}(\mathbf{Z}_{cross}) = \text{Softmax} \left( \frac{\mathbf{Q}_{cross} \mathbf{K}_{cross}^\top}{\sqrt{D}} + \mathbf{M} \right) \mathbf{V}_{cross} \ ,
\end{equation}
the query, key, and value projections are obtained via $\mathbf{Q}_{cross} = \mathbf{Z}_{cross} \mathbf{W}_{cross,Q}$, $\mathbf{K}_{cross} = \mathbf{Z}_{cross} \mathbf{W}_{cross,K}$, $\mathbf{V}_{cross} = \mathbf{Z}_{cross} \mathbf{W}_{cross,V}$. Subsequently, the output of self-attention layer $\mathbf{Z}'_{cross} \in \mathbb{R}^{C \times N_{max} \times D}$ is fed into a standard feed-forward network to get encoder representation $\mathbf{H}_{enc} \in \mathbb{R}^{C \times N_{max} \times D}$:
\begin{equation}
\mathbf{H}_{enc} = \boldsymbol{w}_2 \cdot \text{ReLU}(\mathbf{Z}'_{cross} \boldsymbol{w}_1 + \boldsymbol{b}_1) + \boldsymbol{b}_2 .
\end{equation}
where $\boldsymbol{w}_1$, $\boldsymbol{b}_1$, $\boldsymbol{w}_2$ and $\boldsymbol{b}_2$ are trainable parameters.

\subsection{Prediction Header}
\label{Forecasting}
Given the irregularity present in diverse MTS, the prediction horizon cannot be defined as a fixed window length. Thus, we use a set of queries $\mathbf{Q} = \{q_t\}_{t=T+1}^{T+F}$ corresponding to target prediction timestamps to provide temporal information: 
\begin{equation}
\mathbf{Q}_{emb} = \text{TimeEmb}(\mathbf{Q})\in \mathbb{R}^{F \times D}.
\end{equation}

Since the encoder output $\mathbf{H}_{enc}$ contains comprehensive historical context, we extract the last valid tokens $\mathbf{H}_{last} \in \mathbb{R}^{C \times D}$ and repeat it $F$ times to match the query lengths. The replicated representations are then added with the time embeddings $\mathbf{Q}_{emb}$ to obtain the query representation:
\begin{equation}
\mathbf{H}_{query} = [\text{Repeat}(\mathbf{H}_{last})+ \mathbf{Q}_{emb}] \in \mathbb{R}^{F \times D}. 
\end{equation}
The query representation is then passed through a feed-forward network to get final forecasts for the query timestamps:
\begin{equation}
\hat{\mathbf{Y}} = \boldsymbol{w}_4 \cdot \text{ReLU}(\boldsymbol{w}_3 \cdot \mathbf{H}_{query} + \boldsymbol{b}_3) + \boldsymbol{b}_4, 
\end{equation}
where $\boldsymbol{w}_3$, $\boldsymbol{b}_3$, $\boldsymbol{w}_4$ and $\boldsymbol{b}_4$ are trainable parameters.

\section{Experiments}
\subsection{Experimental Setup}
\textbf{Datasets.}
We conduct experiments on four categories of datasets: 
\textit{Regular MTS datasets} Exchange~\cite{lai2018modeling}, Weather~\cite{wu2021autoformer} and ZafNoo~\cite{poyatos2020global}; 
\textit{Heterogeneous-frequency MTS datasets} constructed by downsampling regular sequences at varying ratios across variables; 
\textit{Missing-value MTS datasets} generated by randomly deleting observations from regular datasets; 
and \textit{Event-driven MTS datasets} Human Activity, PhysioNet2012~\cite{p12} and USHCN~\cite{ushcn}. 
Further details are available in Appendix \ref{Datasets}.

\noindent \textbf{Baselines.}
We compare TiWeaver against 14 baselines, covering both irregular and regular time series forecasting models. 
Irregular models involve \textit{ODE-based methods} NeuralFlows~\cite{nFlows}, GNeuralFlow~\cite{gnflows}, and \textit{graph methods} tPatchGNN~\cite{tpatchgnn}, GraFITi~\cite{GraFITi}, Hi-Patch~\cite{luo2025hipatch}, and \textit{adaptive patching method} APN~\cite{apn} and others mTAND~\cite{mtan}, PrimeNet~\cite{primenet}. 
Regular models include \textit{Transformer-based methods }FEDformer~\cite{fedformer}, PatchTST~\cite{patchtst}, iTransformer~\cite{liu2023itransformer}, and \textit{frequency-based methods} TimesNet~\cite{timesnet}, PDF~\cite{dai2024periodicity}, and \textit{adaptive patching method} TimeMosaic~\cite{mosaic} . 
Additional information is provided in Appendix~\ref{exp details}.

\noindent \textbf{Implementation Details.} 
All experiments are conducted using PyTorch~\cite{paszke2019pytorch} in Python 3.10 and executed on an NVIDIA Tesla-A800 GPU. The training process is guided by the L2 loss, employing the ADAM~\cite{kingma2014adam} optimizer. All experiments are conducted with three random seeds, and the average and standard deviation are reported.

\subsection{Main Results}
To evaluate the performance of TiWeaver, we conduct extensive experiments on 12 datasets, comparing against 14 strong baselines. The complete MAE results are reported in Table~\ref{tab:main_result}, while MSE scores are provided in Appendix~\ref{mse}. As shown in the Table, TiWeaver consistently achieves either the best or second-best performance across all datasets, outperforming the second-best model by an average MAE reduction of over 7\%. In particular, TiWeaver achieves an MAE reduction of 25\% over the suboptimal model on the USHCN dataset. More specifically, we make the following findings. 
On regular and heterogeneous-frequency MTS datasets, fixed-patch methods (e.g., PatchTST) and frequency-based models (e.g., PDF) demonstrate highly competitive performance. Their success indicates that segmenting sequences into equal-length patches or explicitly modeling dominant frequencies can effectively capture stable local patterns and periodicity. However, in datasets characterized by missing observations or event-driven sampling (e.g., Exchange$^*$ and Human Activity), these intrinsic periodicities are frequently disrupted. Consequently, the rigid assumptions of fixed lengths or predefined frequencies become invalid, giving an edge to dynamic partitioning methods like APN and TimeMosaic. Nevertheless, these dynamic baselines still lack universal generalizability across diverse scenarios. To bridge this gap, TiWeaver jointly models the temporal dynamics and fine-grained inter-channel dependencies of MTS. This dual capability allows it to seamlessly align periodic structures when present, while maintaining high robustness against missing values and irregular events, ultimately yielding consistent state-of-the-art improvements across all dataset.

\subsection{Ablation Studies}
To systematically evaluate the contributions of individual components in our model, we conduct extensive ablation studies using the following variants:
(1) \textit{w/o TimeEmb:} Discards the temporal embedding mechanisms across both the inputs and prediction queries.
(2) \textit{w/o GAT:} Substitutes the Graph Attention Network with a standard linear projection for patch-level representation learning.
(3) \textit{w/o G$^2$AT:} Replaces the proposed adaptive patching strategy with a fixed-size patching mechanism.
(4) \textit{w/ KMeans:} Replaces the adaptive patching strategy in G$^2$AT with KMeans clustering.
(5) \textit{w/ DTW:} Replaces the adaptive patching strategy in G$^2$AT with Dynamic Time Warping (DTW).
(6) \textit{w/o FADE:} Removes the Fine-Grained Asynchronous Dependency Extractor, encoding all channels independently via the Transformer encoder without modeling inter-channel dependencies.
(7) \textit{w/ Cross-Attn:} Models inter-channel dependencies using standard cross-attention instead of the proposed method.
(8) \textit{w/ Self-Attn:} Models inter-channel dependencies using standard self-attention.
(9) \textit{w/ Mask-Attn:} Models inter-channel dependencies using masked self-attention.

For simplicity, we select one representative dataset from each type for our experiments. The results are summarized in Table~\ref{ab result}. All ablation variants lead to a noticeable drop in prediction performance, indicating the effectiveness of each module in TiWeaver. Specifically, on Human Activity with a 75\% missing ratio, TiWeaver \textit{w/o TimeEmb} causes the most significant degradation, highlighting its importance in scenarios with highly sparse observations, and the time embedding helps the model recover implicit temporal continuity. On Exchange$^*$ with random missing values, TiWeaver w/o G$^2$AT results in the largest performance drop, demonstrating its ability to adaptively capture diverse dynamics. For both the Heterogeneous-frequency dataset Exchange$^\dag$ and the regular dataset Exchange, TiWeaver w/o FADE exhibits the most substantial decline, suggesting that TiWeaver effectively models fine-grained asynchronous dependencies across channels. 

\begin{table}[h]
\centering
\caption{Effect of variants of TiWeaver in four types of datasets.} 
\resizebox{1\columnwidth}{!}{
\begin{tabular}{c|cc|cc|cc|cc}
\toprule
\multirow{2}{*}{\textbf{Models}} & \multicolumn{2}{c|}{\textbf{Exchange}} & \multicolumn{2}{c|}{\textbf{Exchange$^\dag$}} & \multicolumn{2}{c|}{\textbf{Exchange$^*$}} & \multicolumn{2}{c}{\textbf{Human Activity}} \\
\cmidrule{2-9}
 & MAE & MSE & MAE & MSE & MAE & MSE & MAE & MSE \\
\midrule
w/o TimeEmb & 0.2023 & 0.0846 & 0.1909 & 0.0772 & 0.2156 & 0.0926 & 0.1388 & 0.0570 \\
w/o GAT & 0.1972 & 0.0804 & 0.1909 & 0.0772 & 0.2046 & 0.0852 & 0.1358 & 0.0561 \\
w/o G$^2$AT & 0.2051 & 0.0866 & 0.1906 & 0.0770 & 0.2284 & 0.1042 & 0.1385 & 0.0567 \\
w/ K-Means & 0.2184 & 0.1207 & 0.2153 & 0.0901 & 0.2426 & 0.1152 & 0.1423 & 0.0591 \\
w/ DTW & 0.2217 & 0.1300 & 0.2223 & 0.1004 & 0.2551 & 0.1218 & 0.1714 & 0.0673 \\
w/o FADE & 0.2102 & 0.0907 & 0.1933 & 0.0792 & 0.2071 & 0.0883 & 0.1360 & 0.0575 \\
w/ cross-attn & 0.2054 & 0.0893 & 0.1901 & 0.0764 & 0.2328 & 0.1080 & 0.1360 & 0.0562 \\
w/ self-attn & 0.2067 & 0.0874 & 0.1989 & 0.0817 & 0.2055 & 0.0864 & 0.1356 & 0.562 \\
w/ mask self-attn & 0.2391 & 0.1125 & 0.2040 & 0.0858 & 0.2264 & 0.1010 & 0.1367 & 0.0563 \\
TiWeaver & \textbf{0.1967} & \textbf{0.0795} & \textbf{0.1902} & \textbf{0.0767} & \textbf{0.2022} & \textbf{0.0832} & \textbf{0.1353} & \textbf{0.0554} \\
\bottomrule
\end{tabular}
}
\label{ab result}
\end{table}

\subsection{Case Studies}

\begin{figure}[h]
    \centering
    \includegraphics[width=1\linewidth]{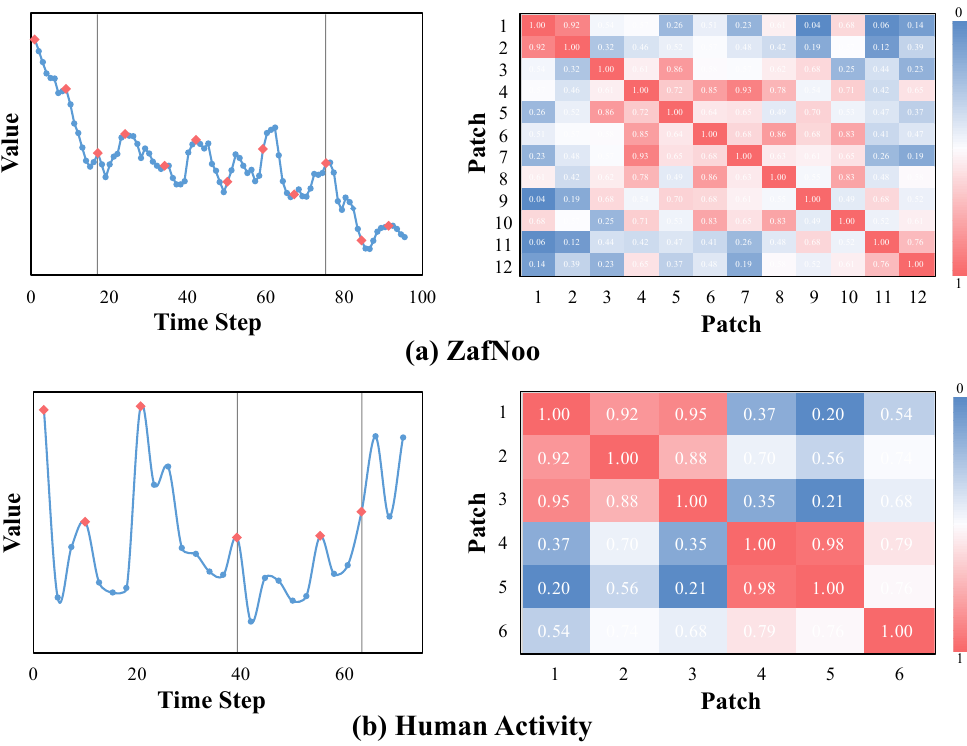}
    \caption{Visualization of adaptive patching. \textit{Left}: Raw time series, where red squares denote the boundaries of adaptively generated patches. \textit{Right}: Pairwise similarity heatmap of these patches (warmer colors for higher similarity)}
    \label{case}
\end{figure}

To intuitively demonstrate the efficacy of adaptive patching, Figure~\ref{case} visualizes the results on two representative datasets: \textit{Exchange} and \textit{Human Activity}, exhibiting pronounced periodic and trend variations. Unlike fixed patching, TiWeaver divides the time series precisely at locations where periodicity or trend changes significantly, avoiding the mixing of heterogeneous dynamics. 

As shown in the left plot of Figure (a), the adaptive boundaries (red squares) naturally isolate three major stages: early sharp decline(Patches 1-2), mid-stage small oscillations (Patches 3-8), and an abrupt drop(Patches 9-12). Notably, in the second stage, TiWeaver further subdivides the sequence into fine-grained local patterns, capturing transient spikes, dips, and brief oscillations. The right heatmap confirms the superiority of this strategy. As major stages with similar periodic or trend dynamics form dark red clusters in the representation space, even when internally finely subdivided into multiple patches, patches with consistent temporal dynamics maintain extremely high similarity, forming prominent red blocks (such as the pairs of Patches 1 \& 2, and Patches 3 \& 5). Furthermore, this robustness extends to highly irregular series. As shown in Figure~\ref{case}(b), the first three patches form a red cluster, as they exhibit a drop followed by a rebound. Notably, the second patch shows a milder decline than the first and third patches, resulting in weaker similarity within this cluster. The fourth and fifth patches form a second cluster with mild fluctuations. The similarity between the two clusters is relatively low. 
The result suggests that adaptive patching strictly aligns with temporal transitions, empowering G$^2$AT to model time series by maximizing intra-patch contextual coherence and promoting inter-patch contextual separability.

\subsection{Parameter Sensitivity Analysis}
In this section, we consider the key parameters min patch size $P_{min}$ and threshold $\tau$ of TiWeaver acting on four types MTS datasets. 

\subsubsection{Effect of min patch size $P_{min}$}
We first investigate the impact of min patch size $P_{min}$, which determines the granularity of initial temporal patches before G$^2$AT. See the polyline in Figure~\ref{fig:patch_threshold}, the choice of $P_{min}$ exhibits different behaviors across datasets. For regular dataset Exchange, an overly small $P_{min}$ (e.g, 1) may cause substantial information loss when TiWeaver initially captures local temporal dependencies, leading to inferior performance. In contrast, on its irregular variants, TiWeaver tends to benefit from starting with a smaller $P_{min}$, which helps better capture disrupted temporal dynamics. Across these three datasets, performance becomes stable around the optimal $P_{min}$ (e.g, 4), indicating the robustness of TiWeaver. However, setting $P_{min}$ too large (e.g., 16, 32) artificially merges discontinuous sequences, which inflates intra-patch heterogeneity and degrades overall performance. For event-driven dataset Human Activity with higher missing rates, TiWeaver is less sensitive to the choice of $P_{min}$. We attribute this to the fact that severe missingness weakens temporal dynamics, thereby reducing the impact of patch granularity.

\subsubsection{Effect of threshold $\tau$}
\begin{figure}[t]
    \centering
    \includegraphics[width=1\linewidth]{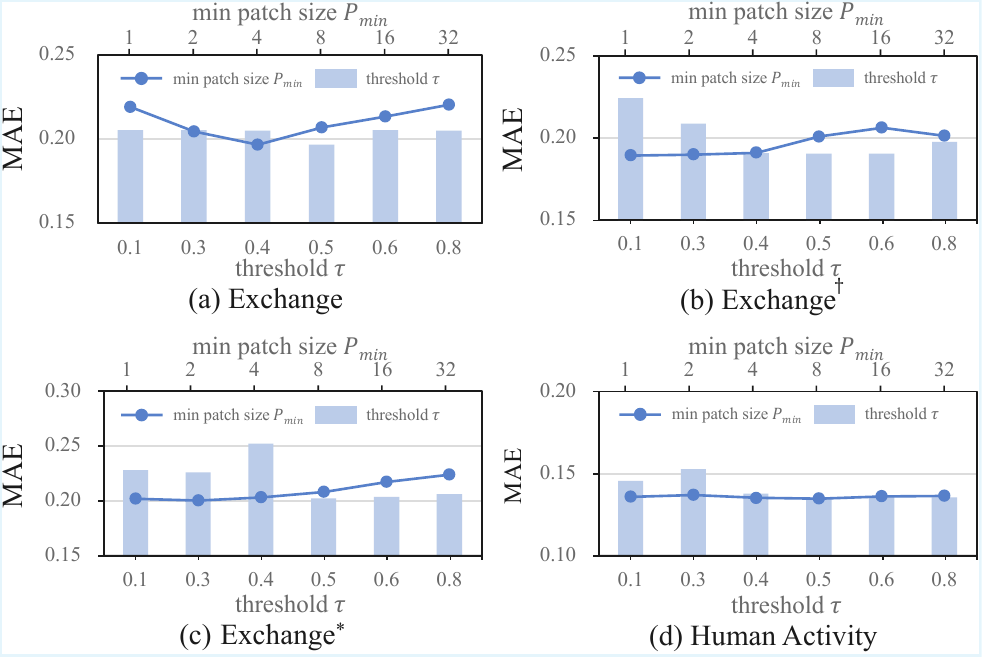}
    \caption{Effect of min patch size $P_{min}$ (top x-axis) and threshold $\tau$ (bottom x-axis).}
    \label{fig:patch_threshold}
  \vspace{-4mm}
\end{figure}

We further examine the threshold $\tau$ of contextual coherence, which controls whether adjacent patches should be treated entirely. 
As shown in the bar plots in Figure~\ref{fig:patch_threshold}(b)-(d), a loose threshold (e.g., 0.1) leads to performance degradation on irregular datasets. This is likely due to overly aggressive merging of patches with distinct representations. 
A moderate threshold (e.g., 0.5) achieves the best results universally, proving the efficacy of contextual coherence-guided adaptive patching. Furthermore, since regular datasets (e.g., Exchange) possess clear periodicity, a well-chosen initial patch size inherently guarantees representational coherence. Therefore, for regular datasets, $P_{min}$ plays a far more critical role than $\tau$, which is in stark contrast to irregular datasets.

\subsection{Efficiency}

\begin{figure}[h]
  \centering
  \includegraphics[width=1\linewidth]{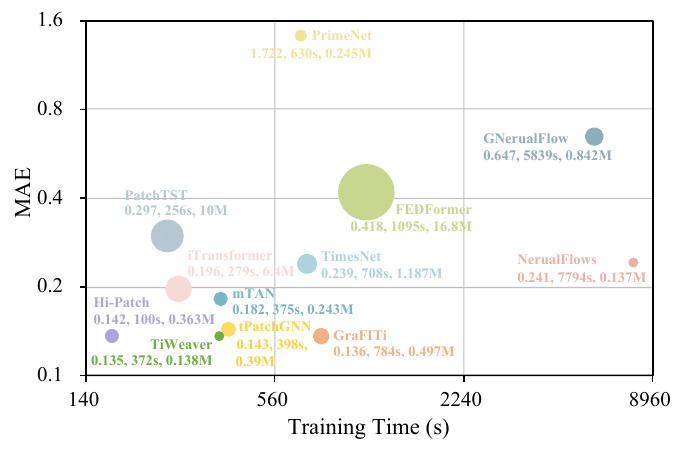}
  \caption{Comparison of model accuracy, training time, and parameter sizes on Human Activity dataset.} 
  \label{fig: time}
\end{figure}

As shown in Figure~\ref{fig: time}, TiWeaver achieves an excellent balance among accuracy, training time, and parameter sizes, validating the effectiveness. Notably, NODE-based models such as NeuralFlows exhibits both high training cost and inferior accuracy, indicating that their potential has not yet been fully realized for MTS forecasting. On the other hand, fixed-length patching methods achieve relatively worse performance, indicating their limitations in modeling irregular time series. In contrast, TiWeaver delivers state-of-the-art accuracy with minimal computational cost, making it a practical solution for modeling diverse MTS.

We analyze the computational complexity to quantitative the efficiency of TiWeaver. 
Let $N$ and $C$ denote the initial patch and channel counts. Although the theoretical worst-case complexity for patching is $\mathcal{O}(CN)$, G$^2$AT guarantees an empirical time complexity of $\mathcal{O}(C\log N)$ by leveraging \textbf{parallel merging} and \textbf{early termination} to avoid linear degradation. As a result, the input sequence is efficiently compressed from $N$ down to $w$ valid patches ($w \ll N$). Operating on this condensed representation, FADE employs a sparse attention strategy that restricts each query to $k$ historical patches. This effectively drops the cross-patch attention overhead to $\mathcal{O}(Cwk)$. Ultimately, the total computational complexity is tightly bounded by $\mathcal{O}(C\log N + Cwk)$. 
As shown in Table~\ref{times}, this theoretical efficiency translates into competitive inference times.

\begin{table}[h]
\centering
\caption{The inference time (in seconds).}
\label{times}
\resizebox{1\linewidth}{!}{
\begin{tabular}{c|c|c|c|c|c|c}
\toprule
\textbf{Time(s)} & \textbf{Exchange} & \textbf{Weather} & \textbf{Zafnoo} & \textbf{Human Activity} & \textbf{USHCN} & \textbf{PhysioNet2012} \\ 
\midrule
NeuralFlows~\cite{nFlows} & 37.4 & 57.0 & 389.2 & 6.0 & 15.4 & 395.0 \\
GNeuralFlow~\cite{gnflows} & 47.5 & 108.2 & 83.1 & 9.2 & 41.8 & 200.8  \\
tPatchGNN~\cite{tpatchgnn} & 2.0 & 15.0 & 6.3 & 2.3 & 14.4 & 9.9 \\
GraFITi~\cite{GraFITi} & 13.0 & 16.9 & 47.2 & 3.3 & 30.1 & 52.8 \\
Hi-Patch~\cite{luo2025hipatch} & 13.3 & 251.8 & 28.2 & 5.0 & 24.7 & 49.3 \\
mTAND~\cite{mtan} & 1.9 & 15.4 & 13.4 & 2.1 & 13.6 & 10.1  \\
PrimeNet~\cite{primenet} & 2.8 & \uline{12.3} & 14.9 & 2.2 & 14.9 & 11.9  \\
APN~\cite{apn} & 4.2 & 18.2 & 7.1 & 2.7 & 16.6 & 16.0 \\
\midrule
FEDformer~\cite{fedformer} & 4.3 & 21.6 & 10.3 & 2.6 & 12.7 & 26.5 \\
PatchTST~\cite{patchtst} & \uline{1.7} & \textbf{10.7} & \textbf{3.5} & 2.0 & 13.3 & 10.2 \\
iTransformer~\cite{liu2023itransformer} & 1.9 & 13.1 & 6.2 & \uline{1.9} & 12.8 & 10.8 \\
TimesNet~\cite{timesnet} & \textbf{1.6} & 22.5 & 6.0 & \textbf{1.8} & \textbf{10.5} & 12.5 \\
PDF~\cite{dai2024periodicity} & 1.8 & 20.3 & 7.6 & - & - & - \\
TimeMosaic~\cite{mosaic} & 9.3 & 18.4 & \uline{3.6} & 2.6 & \uline{11.9} & \uline{9.7} \\
TiWeaver & \textbf{1.6} & 17.1 & 5.4 & \textbf{1.8} & 12.1 & \textbf{8.0} \\
\bottomrule
\end{tabular}
}
\end{table}

\section{Related Work}

\subsection{Temporal Representation and Tokenization}

Time series inherently exhibit both local temporal continuity and long-term dependencies~\cite{wu2025k2vae,wu2025srsnet,wu2026aurora,qiu2026seer}. To effectively model diverse dynamics, it is crucial to tokenize the time series into high contextually coherent patches.
Recent methods such as PatchTST~\cite{patchtst} and TriFormer~\cite{cirstea2022triformer} segment time series into fixed-length patches, treating them as discrete temporal tokens for Transformer-based models. Other methods like TimesNet~\cite{timesnet} and TFMixer~\cite{qiu2026bridging} perform temporal decomposition with FFT or Wavelet Transform to capture multi-scale features. These methods rely on fixed sampling frequency and uniform time interval, rendering them ineffective for irregular time series. 
Consequently, some methods ~\cite{tpatchgnn,luo2025hipatch} operate by partitioning irregular time series into fixed patches, with segmentation determined by time intervals.
Although several recent studies have attempted to flexibly divide complex multivariate time series, they remain constrained by fixed anchor selection~\cite{apn} and a predefined set of patch sizes~\cite{mosaic}. However, these methods exhibit limited flexibility in adapting to diverse dynamics. This limitation highlights the critical need for a unified context-aware tokenization method.

\subsection{Multivariate Correlation}
Effectively modeling inter-channel dependencies is critical for accurate multivariate time series forecasting~\cite{cheng2026cora,qiu2026comprehensive,qiu2026dag,pan2023MagicScaler,wu2024autocts}. While early paradigms either treat channels independently~\cite{patchtst} or assume static topologies, subsequent models like Crossformer~\cite{Crossformer} explicitly capture spatial dynamics via two-stage attention mechanisms. However, such architectures inherently rely on synchronized timestamps and uniform sampling. When confronted with irregular time series, this rigid synchronization inevitably misaligns asynchronous events, severely degrading predictive performance. 
To address structural irregularity, recent graph-based works have been proposed. HiPatch~\cite{luo2025hipatch} builds intra- and inter-patch fully connected graphs for implicit correlation learning, and GraFITi~\cite{GraFITi} aligns channels through timestamp-similarity bipartite graphs. Despite these efforts, their dependency extraction remains structurally coarse-grained.
However, these designs remain inherently coarse-grained, limiting their ability to capture asynchronous interactions.

\section{Conclusion}
In this work, we present TiWeaver, a unified framework for multivariate time series (MTS) forecasting designed to robustly handle diverse data distributions. By introducing the Graph-Guided Adaptive Tokenizer (G$^2$AT) and the Fine-Grained Asynchronous Dependency Extractor (FADE), our model captures complex temporal dynamics and asynchronous inter-channel dependencies in a flexible, context-aware manner. Extensive experiments on 12 real-world datasets demonstrate that TiWeaver consistently achieves state-of-the-art performance across a wide spectrum of scenarios, including regular sampling, heterogeneous frequencies, missing values, and event-driven sequences. Ultimately, these results highlight the effectiveness of our adaptive patching and cross-channel attention designs, paving a promising direction for future research in robust MTS forecasting.

\begin{acks}
This work was partially supported by National Natural Science Foundation of China (62372179)
\end{acks}

\clearpage

\bibliographystyle{ACM-Reference-Format}
\balance
\bibliography{sample-base}


\clearpage

\appendix

\section{Experimental details}
\label{exp details}

\subsection{Datasets}
\label{Datasets}

We assemble a collection of publicly available MTS datasets across six domains, with sampling frequencies ranging from Hz to days and varying lengths.
Summary statistics of all datasets are provided in Table~\ref{tab: Datasets}.
\begin{itemize}
  \item \textbf{Exchange}~\cite{lai2018modeling}: collects daily exchange rates of 8 countries.
  \item \textbf{Weather}~\cite{wu2021autoformer}: collects 21 meteorological indicators, including temperature and barometric pressure, for Germany in 2020, recorded every 10 minutes.
  \item \textbf{ZafNoo}~\cite{poyatos2020global}: includes sap flow measurements alongside environmental observations drawn from the Sapflux Data Project.
  \item \textbf{Human Activity}: contains biomechanics data of 3D positional variables, sampled at 1 ms resolution.
  \item \textbf{USHCN}~\cite{ushcn}: provides climate records from U.S. weather stations from 1996 to 2000.
  \item \textbf{PhysioNet2012}~\cite{p12}: comprises clinical measurements taken during the first 48 hours of ICU stay in 1-hour intervals.
\end{itemize}

Beyond these real‐world collections, we construct two types of synthetic MTS datasets:
\begin{enumerate}
  \item \textbf{Heterogeneous‐frequency MTS}: produced by downsampling each channel at different ratios.
  \item \textbf{Missing‐value MTS}: derived from the regular datasets introduced above by randomly deleting 20\% of all observations across all channels.
\end{enumerate}

\begin{table*}[t]
\centering
\small
\caption{The statistics of evaluation datasets.}
\label{tab: Datasets}
\resizebox{1.0\linewidth}{!}{
\begin{tabular}{c|c|c|c|c|c|c|c|c}
\toprule
\textbf{Description} & \textbf{Samples} & \textbf{Variables} & \textbf{Missing Ratio} & \textbf{Domain} & \textbf{Frequency} & \textbf{Split Ratio} & \textbf{Observation} & \textbf{Prediction} \\
\midrule
Exchange & 7588 & 8 & 0\% & Economic & 1 day & 7:1:2 & 96 & 96 \\
Weather & 52696 & 21 & 0\% & Environment & 10 mins & 7:1:2 & 96 & 96 \\
ZafNoo & 19255 & 11 & 0\% & Nature & 30 mins & 7:1:2 & 96 & 96 \\
Human Activity & 5400 & 12 & 75\% & Human Activity & 50Hz & 6:2:2 & 3000ms & 1000ms \\
USHCN & 1114 & 5 & 77.90\% & Meteorology & 1 day & 6:2:2 & 24months & 1month \\
PhysioNet2012 & 12000 & 36 & 85.70\% & Medicine & 1 hour & 6:2:2 & 24h & 24h \\
\bottomrule
\end{tabular}
}
\end{table*}

\subsection{Baselines}
\label{Baselines}

We evaluate TiWeaver against 14 baselines.

\begin{itemize}
  \item \textbf{NeuralFlows}~\cite{nFlows} parameterizes the solution of an ODE with a neural network, enabling continuous-time modeling without costly solver iterations.
  \item \textbf{GNeuralFlow}~\cite{gnflows} augments NeuralFlows with GNN to capture inter-channel interactions in irregular MTS.
  \item \textbf{tPatchGNN}~\cite{tpatchgnn} divides irregular MTS into patches and applies a GNN over these patches to jointly capture local and global dependencies.
  \item \textbf{GraFITi}~\cite{GraFITi} represents observations and channels as nodes in a bipartite graph, performing message passing across temporal and channel dimensions.
  \item \textbf{Hi-Patch}~\cite{luo2025hipatch} hierarchically constructs intra-patch and inter-patch graphs, modeling fine-grained and multi-scale dependencies for irregular MTS.
  \item \textbf{mTAND}~\cite{mtan} embeds continuous timestamps into a learned representation and employs self-attention to produce fixed-length features for irregular MTS.
  \item \textbf{PrimeNet}~\cite{primenet} uses a pre-training regime on irregular MTS to learn generalizable representations before downstream forecasting tasks.
  \item \textbf{APN}~\cite{apn} introduces an aggregation-based paradigm for adaptive patching.
  \item \textbf{FEDformer}~\cite{fedformer} decomposes signals in the frequency domain by randomly selecting a subset of Fourier bases, representing both low- and high-frequency components for efficient forecasting.
  \item \textbf{PatchTST}~\cite{patchtst} segments long time series into fixed-size patches and treats each channel independently, learning patch-level dependencies to improve long-term forecasting.
  \item \textbf{iTransformer}~\cite{liu2023itransformer}applies a Transformer architecture on inverted channel dimensions, allowing attention and feed-forward layers to directly model inter-channel correlations.
  \item \textbf{TimesNet}~\cite{timesnet} reformulates 1D series into 2D to uncover multiple periodicity via a lightweight inception block.
  \item \textbf{PDF}~\cite{dai2024periodicity} maps 1D series to 2D and performs multi-period frequency-domain decoupling to jointly capture short-term fluctuations and long-term trends.
  \item \textbf{TimeMosaic}~\cite{mosaic} employs adaptive patch embedding to adjust granularity according to local information density.
\end{itemize}

\section{Theoretical Analysis}
\label{theoretical analysis}
To substantiate the claim that G$^2$AT models temporal dynamics more effectively than fixed patching, by maximizing intra-patch contextual coherence and promoting inter-patch contextual separability, we provide a theoretical analysis here. 

Given a time series $\mathbf{X} = \{x_i\}_{i=1}^T$, complex temporal dynamics can be approximated by a set of piecewise-stationary patches $\{\mathbf{X}_k\}_{k=1}^K$. Following the Minimum Description Length (MDL) principle~\cite{mdl}, we select patch boundaries that best trade-off the number and the intra-coherence of patches, yielding an adaptive tokenization:
\begin{equation}
MDL = \lambda K + \sum_{k=1}^K L(\mathbf{X}_k),
\end{equation}
where $K$ is the total number of patches, and $L(\mathbf{X}_k)$ measures the intra-patch heterogeneity, which means a smaller $L(\mathbf{X}_k)$ indicates higher contextual coherence.

G$^2$AT performs adaptive patching in a bottom-up process. For each pair of adjacent patches $\mathbf{X}_i$ and $\mathbf{X}_{i+1}$, G$^2$AT evaluates their representation consistency and temporal density $Sim(\mathbf{X}_i, \mathbf{X}_{i+1}) * \delta_i$. The candidate contextually coherent patch $\mathbf{X}_m = \mathbf{X}_i \cup \mathbf{X}_{i+1}$ is generated when the following conditions are satisfied:
\begin{equation}
Sim(\mathbf{X}_i,\mathbf{X}_{i+1}) > \tau \ \ \& \ \
Sim(\mathbf{X}_m,\mathbf{X}_i) > \tau \ \ \& \ \
Sim(\mathbf{X}_m,\mathbf{X}_{i+1}) > \tau.
\label{eq9}
\end{equation}
where $\tau$ is a predefined coherent threshold. This procedure implies that $\mathbf{X}_i$ and $\mathbf{X}_{i+1}$ are highly contextually coherent and the resulting $\mathbf{X}_m$ preserves the same contextual information. This indicates that treating $\mathbf{X}_i$ and $\mathbf{X}_{i+1}$ as an entire sequence does not increase the heterogeneity of the representation:
\begin{equation}
    L(\mathbf{X}_{i}) \approx L(\mathbf{X}_{i+1}) \approx L(\mathbf{X}_{m}),
\end{equation}

This strategy can be viewed as an approximation to the MDL objective, aiming to minimize the change in MDL ($\Delta$MDL):

\begin{equation}
\begin{aligned}
\Delta MDL
&= MDL(\mathbf{X}_{m})-((MDL(\mathbf{X}_{i}) + MDL(\mathbf{X}_{i+1})) \\
&= -\lambda + L(\mathbf{X}_m) - L(\mathbf{X}_i) - L(\mathbf{X}_{i+1}) \\
&= -\lambda - L(\mathbf{X}_m).
\end{aligned}
\end{equation}
Since $\Delta MDL$ is always less than 0, TiWeaver partitions the time series into a smaller number of discretized tokens, in accordance with the MDL principle.

\section{Additional Experiments}
We report the MSE results in Table~\ref{mse}.

\begin{table*}[t]
\small
\centering
\caption{Overall performance evaluated by MSE (mean ± std). The best-performing results are highlighted in bold, the second-best results are highlighted in underline.}
\resizebox{1\linewidth}{!}{
\begin{tabular}{c|ccc|ccc}
\toprule

\multirow{2}{*}{\textbf{Datasets}} & \multicolumn{3}{c|}{\textbf{Regular MTS}} & \multicolumn{3}{c}{\textbf{Heterogeneous-frequency MTS}} \\ \cmidrule{2-7} 
 & \textbf{Exchange} & \textbf{Weather} & \textbf{ZafNoo} & \textbf{Exchange$^\dag$} & \textbf{Weather$^\dag$} & \textbf{ZafNoo$^\dag$} \\ \midrule

NeuralFlows~\cite{nFlows} & 1.6251$\pm$0.3652 & 0.1764$\pm$0.0090 & 0.5715$\pm$0.0303 & 0.9845$\pm$0.0472 & 0.1590$\pm$0.0072 & 0.7544$\pm$0.0058 \\
GNeuralFlow~\cite{gnflows} & 2.0110$\pm$0.8804 & 0.5762$\pm$0.1997 & 0.8605$\pm$0.0976 & 1.6406$\pm$0.1445 & 0.6490$\pm$0.2377 & 0.9228$\pm$0.1130 \\
tPatchGNN~\cite{tpatchgnn} & 0.3507$\pm$0.1147 & 0.1695$\pm$0.0026 & 0.4946$\pm$0.0222 & 0.2218$\pm$0.0263 & 0.1628$\pm$0.0011 & 0.7114$\pm$0.0122 \\
GraFITi~\cite{GraFITi} & 0.3480$\pm$0.0543 & 0.1833$\pm$0.0117 & 0.4878$\pm$0.0073 & 0.1992$\pm$0.0627 & 0.1573$\pm$0.0075 & 0.7272$\pm$0.0048 \\
Hi-Patch~\cite{luo2025hipatch} & 0.4945$\pm$0.1649 & 0.2140$\pm$0.0067 & 0.5549$\pm$0.0270 & 0.1943$\pm$0.0124 & 0.1750$\pm$0.0208 & 0.7443$\pm$0.0455 \\ 
mTAND~\cite{mtan} & 0.9690$\pm$0.1233 & 0.2543$\pm$0.0088 & 0.5532$\pm$0.0230 & 0.8896$\pm$0.1021 & 0.1780$\pm$0.0291 & 0.5204$\pm$0.0265 \\
PrimeNet~\cite{primenet} & 3.1088$\pm$0.0201 & 0.6365$\pm$0.0001 & 0.8910$\pm$0.0015 & 2.8975$\pm$0.0014 & 0.6249$\pm$0.0001 & 0.8870$\pm$0.0000 \\  
APN~\cite{apn} & 0.2109$\pm$0.1002 & 0.1971$\pm$0.0200 & 0.7312$\pm$0.0076 & 0.1320$\pm$0.0513 & 0.1610$\pm$0.0068 & 0.7195$\pm$0.0009 \\ \midrule
FEDformer~\cite{fedformer} & 0.1587$\pm$0.0092 & 0.2274$\pm$0.0062 & 0.7910$\pm$0.1167 & 1.1094$\pm$0.8543 & 0.2761$\pm$0.0674 & 0.7523$\pm$0.1752 \\
PatchTST~\cite{patchtst} & \uline{0.0845$\pm$0.0006} & 0.1701$\pm$0.0014 & 0.4724$\pm$0.0008 & 0.0932$\pm$0.0018 & 0.3357$\pm$0.0008 & \uline{0.4630$\pm$0.0004} \\
iTransformer~\cite{liu2023itransformer} & 0.1044$\pm$0.0042 & 0.2069$\pm$0.0016 & 0.4832$\pm$0.0029 & 0.1433$\pm$0.0145 & 0.2040$\pm$0.0061 & 0.7503$\pm$0.0011 \\
TimesNet~\cite{timesnet} & 0.1218$\pm$0.0090 & 0.1806$\pm$0.0074 & 0.6508$\pm$0.1371 & 0.2283$\pm$0.0170 & 0.1642$\pm$0.0105 & 0.6344$\pm$0.1657 \\
PDF~\cite{dai2024periodicity} & 0.0873$\pm$0.0022 & \textbf{0.1639$\pm$0.0004} & \textbf{0.4616$\pm$0.0050} & 0.0961$\pm$0.0081 & \uline{0.1503$\pm$0.0021} & 0.5535$\pm$0.0003 \\  
TimeMosaic~\cite{mosaic} & 0.0923$\pm$0.0227 & 0.1820$\pm$0.0003 & 0.4674$\pm$0.0591 & \uline{0.0879$\pm$0.0359} & 0.1940$\pm$0.0080 & 0.4753$\pm$0.1209 \\ \midrule
\textbf{TiWeaver} & \textbf{0.0828$\pm$0.0029} & \uline{0.1644$\pm$0.0081} & \uline{0.4648$\pm$0.0025} & \textbf{0.0778$\pm$0.0010} & \textbf{0.1477$\pm$0.0007} & \textbf{0.4532$\pm$0.0047} \\ 

\bottomrule
\toprule

\multirow{2}{*}{\textbf{Datasets}} & \multicolumn{3}{c|}{\textbf{Missing-value MTS}} & \multicolumn{3}{c}{\textbf{Event-driven MTS}} \\ 
\cmidrule{2-7}
 & \textbf{Exchange$^*$} & \textbf{Weather$^*$}  & \textbf{ZafNoo$^*$} & \textbf{Human Activity} & \textbf{USHCN} & \textbf{PhysioNet2012} \\
\midrule

NeuralFlows~\cite{nFlows} & 1.1181$\pm$0.2580 & 0.1847$\pm$0.0015 & 0.7698$\pm$0.0052 & 0.2375$\pm$0.0150 & 0.5103$\pm$0.0033 & 0.4750$\pm$0.0030 \\
GNeuralFlow~\cite{gnflows} & 2.1252$\pm$0.3353 & 0.6218$\pm$0.0832 & 0.9130$\pm$0.0603 & 0.3529$\pm$0.0099 & 0.7280$\pm$0.1940 & 0.7727$\pm$0.0491 \\
tPatchGNN~\cite{tpatchgnn} & 0.2456$\pm$0.0083 & 0.1717$\pm$0.0017 & 0.7278$\pm$0.0077 & 0.0589$\pm$0.0011 & 0.4774$\pm$0.0218 & 0.5520$\pm$0.2078 \\
GraFITi~\cite{GraFITi} & 0.3360$\pm$0.0600 & 0.1778$\pm$0.0097 & 0.7328$\pm$0.0017 & 0.0766$\pm$0.0022 & \uline{0.4758$\pm$0.0124} & 0.5077$\pm$0.0004 \\
Hi-Patch~\cite{luo2025hipatch} & 0.3525$\pm$0.2221 & 0.1931$\pm$0.0082 & 0.7638$\pm$0.0189 & 0.0575$\pm$0.0004 & 0.4920$\pm$0.0060 & \textbf{0.4418$\pm$0.0005} \\ 
 mTAND~\cite{mtan} & 0.8539$\pm$0.1211 & 0.1984$\pm$0.0065 & 0.5537$\pm$0.0485 & 0.1104$\pm$0.0072 & 1.1993$\pm$0.1903 & 0.4680$\pm$0.0016 \\
PrimeNet~\cite{primenet} & 3.1073$\pm$0.0012 & 0.6371$\pm$0.0001 & 0.8911$\pm$0.0000 & 4.4244$\pm$0.0014 & 0.8223$\pm$0.0000 & 0.7928$\pm$0.0000 \\  
APN~\cite{apn} & 0.1785$\pm$0.0714 & 0.1758$\pm$0.0085 & 0.7340$\pm$0.0004 & \textbf{0.0563$\pm$0.0104} & 0.5553$\pm$0.0311 & 0.4738$\pm$0.0077 \\ \midrule
FEDformer~\cite{fedformer} & 0.3115$\pm$0.0859 & 0.2771$\pm$0.0228 & 0.7633$\pm$0.1690 & 0.2511$\pm$0.0063 & 0.8389$\pm$0.2566 & 0.5019$\pm$0.0001 \\ 
PatchTST~\cite{patchtst} & 0.1849$\pm$0.0062 & 0.1834$\pm$0.0015 & \uline{0.5243$\pm$0.0007} & 0.0797$\pm$0.0003 & 0.6002$\pm$0.0898 & 0.4952$\pm$0.0390 \\
iTransformer~\cite{liu2023itransformer} & 0.1899$\pm$0.0112 & 0.2187$\pm$0.0068 & 0.7760$\pm$0.0015 & 0.1326$\pm$0.0110 & 0.6452$\pm$0.0266 & 0.5058$\pm$0.0011 \\
TimesNet~\cite{timesnet} & 0.2373$\pm$0.0688 & 0.1857$\pm$0.0011 & 0.6446$\pm$0.1462 & 0.1337$\pm$0.0009 & 0.4947$\pm$0.0082 & 0.4981$\pm$0.0018 \\
PDF~\cite{dai2024periodicity} & 0.1497$\pm$0.0133 & \uline{0.1705$\pm$0.0011} & 0.6346$\pm$0.0054 & - & - & - \\  
TimeMosaic~\cite{mosaic} & \uline{0.1186$\pm$0.0362} & 0.2741$\pm$0.0039 & 0.5557$\pm$0.0006 & 0.1515$\pm$0.0115 & 0.5338$\pm$0.0200 & 0.5051$\pm$0.0217 \\ \midrule
\textbf{TiWeaver} & \textbf{0.0893$\pm$0.0053} & \textbf{0.1651$\pm$0.0054} & \textbf{0.5125$\pm$0.0034} & \uline{0.0569$\pm$0.0013} & \textbf{0.4634$\pm$0.0002} & \uline{0.4647$\pm$0.0063} \\ 
\bottomrule
\end{tabular}
}

\begin{tablenotes}
\fontsize{8}{2}
\selectfont
\item \textit{\textbf{Note:}} PDF impose a rigid constraint between the input sequence length and the patch size, which are unable to handle event-driven datasets with variable-length inputs, as demonstrated in “–”.
\end{tablenotes}

\label{mse}
\end{table*}

\end{document}